\documentclass[10pt,twocolumn,letterpaper]{article}

\usepackage{cvpr}
\usepackage{times}
\usepackage{epsfig}
\usepackage{graphicx}
\usepackage{amsmath}
\usepackage{amssymb}
\usepackage{color}
\usepackage{booktabs}
\usepackage{subcaption}

\usepackage[breaklinks=true,bookmarks=false]{hyperref}

\cvprfinalcopy 

\def\cvprPaperID{6265} 

\ifcvprfinal\fi
\begin{document}

\title{CenterMask: single shot instance segmentation with point representation}

\author{Yuqing Wang\qquad Zhaoliang Xu\qquad Hao Shen\qquad Baoshan Cheng \qquad Lirong Yang
\\Meituan Dianping Group
\\{\tt\small \{wangyuqing06,xuzhaoliang,shenhao04,chengbaoshan02,yanglirong\}@meituan.com}}

\maketitle
\thispagestyle{empty}

\begin{abstract}
  In this paper, we propose a single-shot instance segmentation method, which is simple, fast and accurate. 
There are two main challenges for one-stage instance segmentation: object instances differentiation and pixel-wise feature alignment.
Accordingly, we decompose the instance segmentation into two parallel subtasks: Local Shape prediction that separates instances even in overlapping conditions, and Global Saliency generation that segments the whole image in a pixel-to-pixel manner.
The outputs of the two branches are assembled to form the final instance masks. 
To realize that, the local shape information is adopted from the representation of object center points. 
Totally trained from scratch and without any bells and whistles, the proposed CenterMask achieves 34.5 mask AP with a speed of 12.3 fps, using a single-model with single-scale training/testing on the challenging COCO dataset.
The accuracy is higher than all other one-stage instance segmentation methods except the 5 times slower TensorMask, which shows the effectiveness of CenterMask. 
Besides, our method can be easily embedded to other one-stage object detectors
such as FCOS and performs well, showing the generalization of CenterMask.
\end{abstract}

\section{Introduction}

Instance segmentation~\cite{hariharan2014simultaneous} is a fundamental and challenging computer vision task, which requires to locate, classify, and segment each instance in the image. Therefore, it has both the characters of object detection and semantic segmentation. State-of-the-art instance segmentation methods \cite{he2017mask,liu2018path,huang2019mask} are mostly built on the advances of two-stage object detectors \cite{girshick2014rich,girshick2015fast,ren2015faster}. Despite the popular trend of one-stage object detection \cite{huang2015densebox,redmon2016you,liu2016ssd,law2018cornernet,tian2019fcos,zhou2019objects}, only a few works\cite{bolya-iccv2019,chen2019tensormask,xie2019polarmask,fu2019retinamask} are focusing on one-stage instance segmentation. \textit{In this work, we aim to design a simple one-stage and anchor-box free instance segmentation model}.

\begin{figure}[t]
\includegraphics[width=\linewidth]{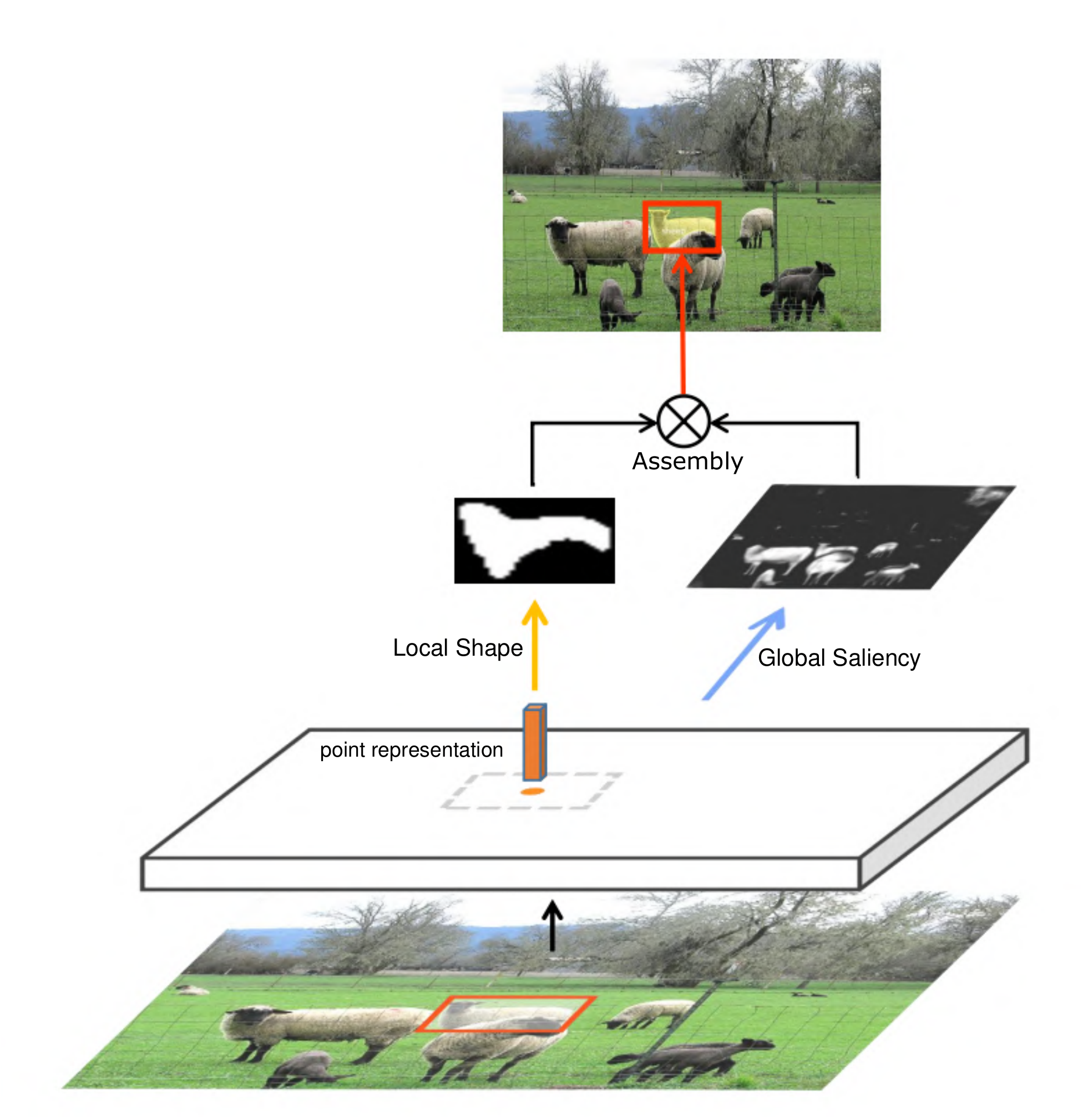}
 \vspace{-0.7cm}
  \caption{\textbf{Illustration of CenterMask.} The Local Shape branch separates objects locally and the Global Saliency Map realizes pixel-wise segmentation of the whole image. Then the coarse but instance-aware local shape and the precise but instance-unaware global saliency map are assembled to form the final instance mask. }
\label{fig:cover}
\vspace{-0.4cm}
\end{figure}

\begin{figure*}[t]
\includegraphics[width=\linewidth]{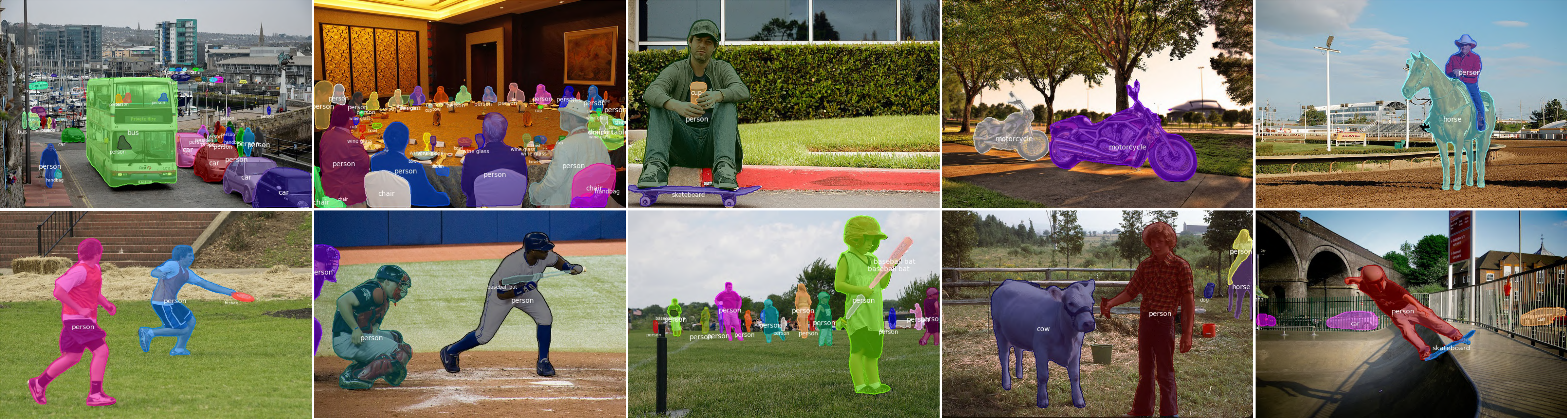}
\vspace{-0.6cm}
  \caption{\textbf{Results of CenterMask on COCO test set images}. These results are based on Hourglass-104 backbone, achieving a mask AP of 34.5 and running at 12.3 fps. Our method differentiates objects well in overlapping conditions with precise masks.}
\label{fig:test}
\vspace{-0.4cm}
\end{figure*}

Instance segmentation is much harder than object detection because the shapes of instances are more flexible than the two-dimensional bounding boxes. 
There are two main challenges for one-stage instance segmentation: (1) how to differentiate object instances, especially when they are in the same category. Some methods \cite{dai2016instance,bolya-iccv2019} extract the global features of the image firstly then post-process them to separate different instances, but these methods struggle when objects overlap. (2) how to preserve pixel-wise location information. State-of-the-art methods represent masks as structured 4D tensors \cite{chen2019tensormask} or contour of fixed points \cite{xie2019polarmask}, but still facing the pixel misalignment problem, which makes the masks coarse at the boundary. TensorMask \cite{chen2019tensormask} designs complex pixel align operations to fix this problem, which makes the network even slower than the two-stage counterparts.

To address these issues, we propose to break up the mask representation into two parallel components: (1) a Local Shape representation that predicts a coarse mask for each local area, which can separate different instances automatically. (2) a Global Saliency Map that segment the whole image, which can provide saliency details, and realize pixel-wise alignment.
To realize that, the local shape information is extracted from the point representation at object centers. Modeling object as its center point is motivated by the one-stage CenterNet \cite{zhou2019objects} detector, thus we call our method CenterMask.

The illustration of the proposed CenterMask is shown in Figure~\ref{fig:cover}. Given an input image, the object center point locations are predicted following a keypoint estimation pipeline. Then the feature representation at the center point is extracted to form the local shape, which is represented by a coarse mask that separates the object from close ones. In the meantime, the fully convolutional backbone produces a global saliency map of the whole image, which separates the foreground from the background at pixel level. Finally, the coarse but instance-aware local shapes and the precise but instance-unaware global saliency map are assembled to form the final instance masks.

To demonstrate the robustness of CenterMask and analyze the effects of its core factors, extensive ablation experiments are conducted and the performance of multiple basic instantiations are compared. Visualization shows that the CenterMask with only Local Shape branch can separate objects well, and the model with only Global Seliency branch performs good enough in objects-non-overlapping situations. In complex and objects-overlapping situations, combination of these two branches differentiates instances and realizes pixel-wise segmentation simultaneously. Results of CenterMask on COCO \cite{lin2014microsoft} test set images are shown in Figure~\ref{fig:test}.

In summary, the main contributions of this paper are as follows:
\begin{itemize}
\item An anchor-box free and one-stage instance segmentation method is proposed, which is simple, fast and accurate. Totally trained from scratch, the proposed CenterMask achieves 34.5 mask AP with a speed of 12.3 fps on the challenging COCO, showing the good speed-accuracy trade-off. Besides, the method can be easily embedded to other one-stage object detectors such as FCOS\cite{tian2019fcos} and performs well, showing the generalization of CenterMask.

\item The Local Shape representation of object masks is proposed to differentiate instances in the anchor-box free condition. Using the representation of object center points, the Local Shape branch predicts coarse masks and separate objects effectively even in the overlapping situations.

\item The Global Saliency Map is proposed to realize pixel-wise feature alignment naturally. Different from previous feature align operations for instance segmentation, this module is simpler, faster, and more precise.
The Global Saliency generation acts similar to semantic segmentation \cite{long2015fully}, and hope this work can motivate one-stage panoptic segmentation \cite{Kirillov_2019_CVPR} in the future.

\end{itemize}

\begin{figure*}[!t]
\centering
\includegraphics[width=.98\linewidth]{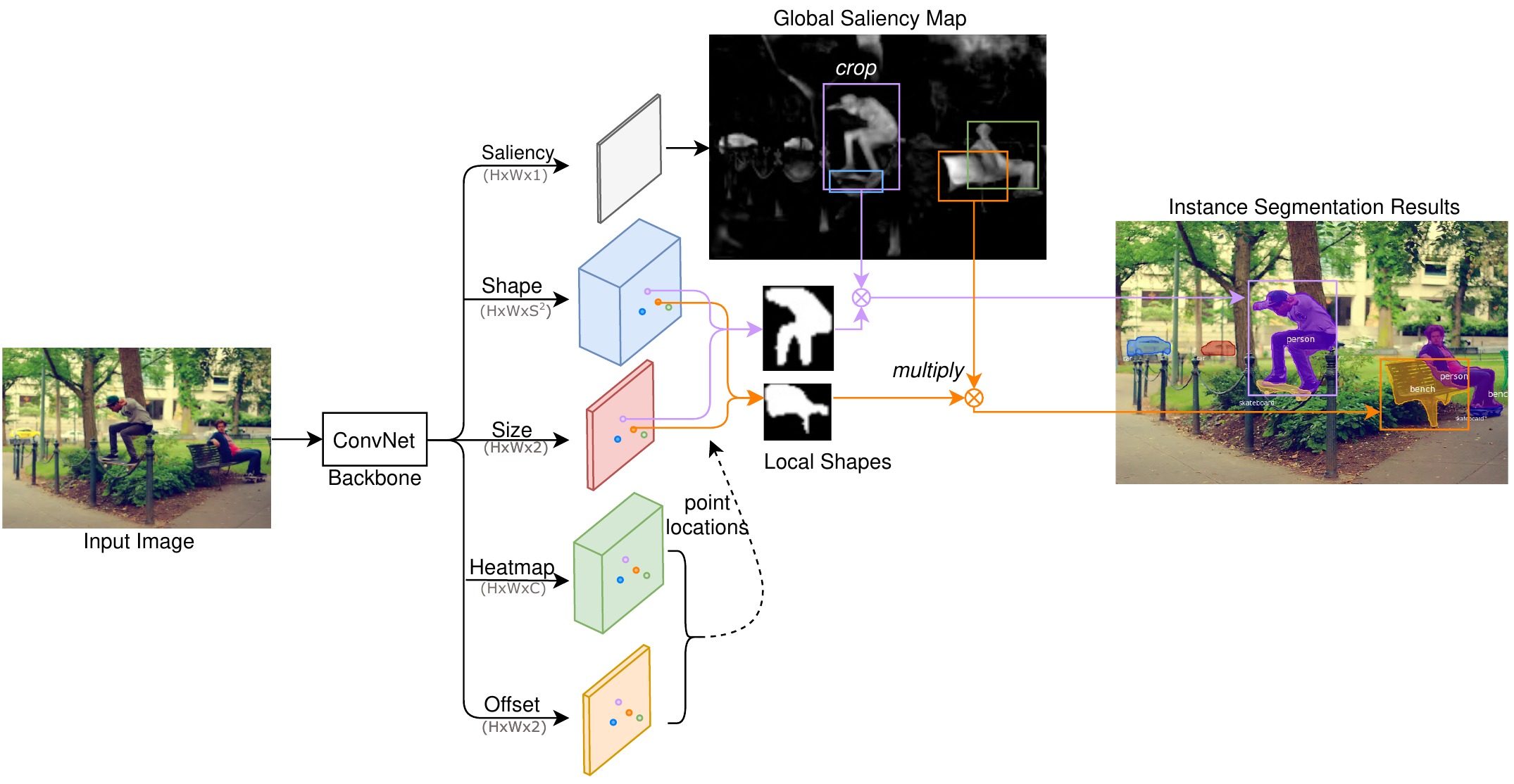}
\vspace{-0.4cm}

\caption{\textbf{Overall pipeline of CenterMask.} There are five heads after the backbone network. The outputs of the heads are with the same height (H) and width (W) but different channels. C is the number of categories, and $\rm S^2$ is the size of shape vector. The Heatmap and Offset heads predict the center point locations. The Shape and Size heads predict the Local Shapes at the corresponding locations. The Saliency head predicts a Global Saliency Map. The Local Shape and cropped Saliency Map are multiplied to form the final mask for each instance. For visualization convenience, the whole segmentation pipeline for only two instances is shown in the figure, and the Global Saliency Map is visualized in the class-agnostic form.
}
\label{fig:network}
\vspace{-0.3cm}
\end{figure*}

\section{Related Work}

\textbf{Two-stage Instance Segmentation:} Two-stage instance segmentation method often follows the \textit{detect-then-segment} paradigm, which first performs bounding box detection and then classifies the pixels in the bounding box area to obtain the final mask. 
Mask R-CNN \cite{he2017mask} extends the successful Faster R-CNN \cite{ren2015faster} detector by adding a mask segmentation branch on each Region of Interest area. To preserve the exact spatial locations, it introduces the RoIAlign module to fix the pixel misalignment problem. 
PANet \cite{liu2018path} aims to improve the information propagation of Mask R-CNN by introducing bottom-up path augmentation, adaptive feature pooling, and fully-connected fusion.
Mask Scoring R-CNN \cite{huang2019mask} proposes a mask scoring module instead of the classification score to evaluate the mask, which can improve the quality of the segmented mask.

Although two-stage instance segmentation methods achieve state-of-the-art performance, these models are often complicated and slow. Advances of one-stage object detection motivate us to develop faster and simpler one-stage instance segmentation methods.

\textbf{One-stage Instance Segmentation:} State-of-the-art one-stage instance segmentation methods can be roughly divided into two categories: \textit{global-area-based} and \textit{local-area-based} approaches. 
 \textit{Global-area-based} methods first generate intermediate and shared feature maps based on the whole image, then assemble the extracted features to form the final masks for each instance.

InstanceFCN \cite{dai2016instance} utilizes FCN \cite{long2015fully} to generate multiple instance-sensitive score maps which contain the relative positions to objects instances, then applies an assembling module to output object instances. YOLACT \cite{bolya-iccv2019} generates multiple prototype masks of the global image, then utilizes per-instance mask coefficients to produce the instance level mask. \textit{Global-area-based} methods can maintain the pixel-to-pixel alignment which makes masks precise, but performs worse when objects overlap.
In contrast to these methods, \textit{local-area-based} methods output instance masks on each local region directly.
PolarMask \cite{xie2019polarmask} represents mask in its contour form and utilizes rays from the center to describe the contour, but the polygon surrounded by the contour can not depict the mask precisely and can not describe objects that have holes in the center. 
TensorMask \cite{chen2019tensormask} utilizes structured 4D tensors to represent masks over a spatial domain, it also introduces aligned representation and tensor bipyramid to recover spatial details, but these align operations make the network even slower than the two-stage Mask R-CNN \cite{he2017mask}. 

Different from the above approaches, CenterMask contains both a Global Saliency generation branch and a Local Shape prediction branch, and integrates them to preserve pixel alignment and separate objects simultaneously.

\section{CenterMask}

The goal of this paper is to build a one-stage instance segmentation method. One-stage means that there is no pre-defined Region-of-Interests (RoIs) for mask prediction, which requires to locate, classify, and segment objects simultaneously. To realize that, we break the instance segmentation into two simple and parallel sub-tasks, and assemble the results of them to form the final masks. The first branch predicts coarse shape from the center point representation of each object, which can constrain the local area for each object and differentiate instances naturally. The second branch predicts a saliency map of the whole image, which realizes precise segmentation and preserves exact spatial locations. In the end, the mask for each instance is constructed by multiplying the outputs of the two branches.

\subsection{Local Shape Prediction}\label{sec:local}

\begin{figure}[!t]
\includegraphics[width=\linewidth]{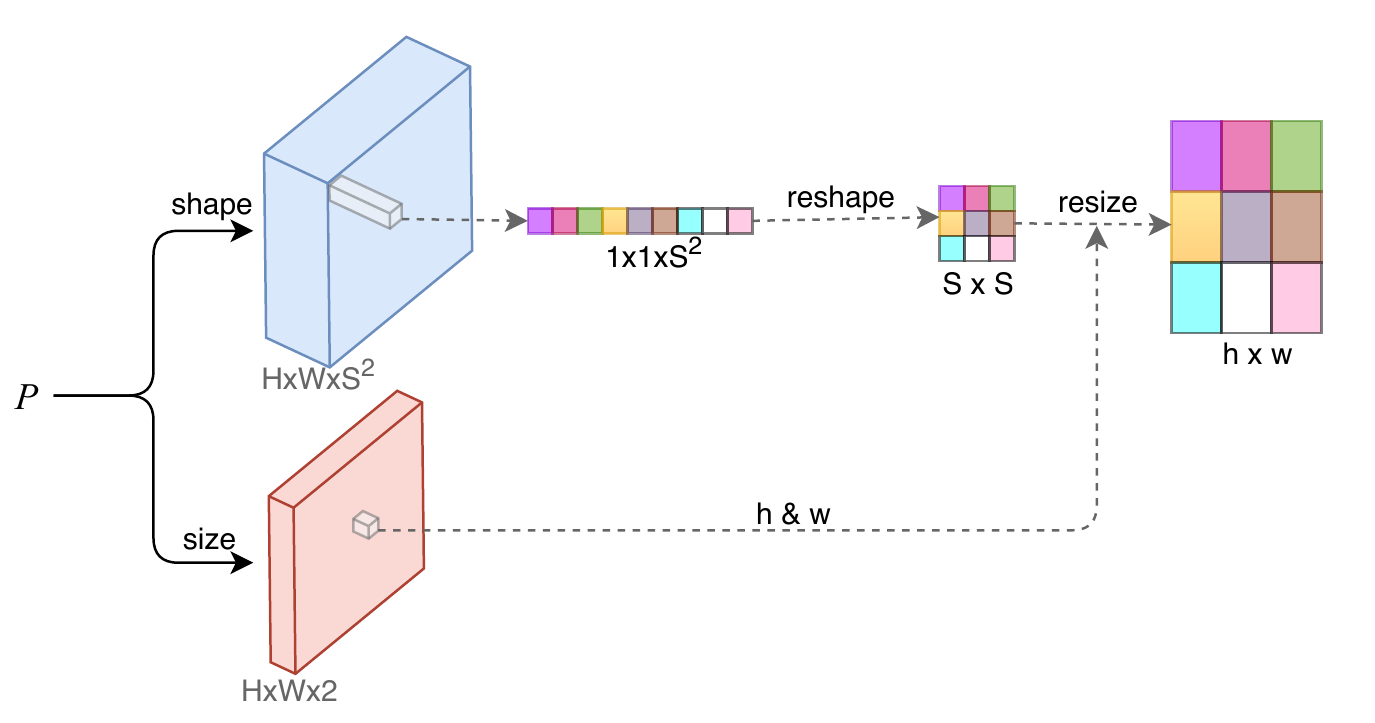}
\vspace{-0.7cm}
  \caption{\textbf{Architecture of the shape head and size head for Local Shape prediction.} $P$ represents the feature maps extracted by the backbone network. H and W represents the height and width of the head outputs. The channel size of the shape head is SxS, and the channels of the size head is 2, with h and w being the predicted height and width for the object at the point. }
\label{fig:shape_branch}
\vspace{-0.3cm}
\end{figure}

To differentiate instances at different locations, we choose to model the masks from their center points. The center point is defined as the center of the surrounding bounding box for each object. A natural thought is to represent it by the extracted image feature at the center point location, but a fixed-size image feature can not represent masks in various sizes. To address this issue, we decompose the object mask into two components: the mask size and the mask shape. The size for each mask can be represented by the object height and width, and the shape can be described by a 2D binary array of fixed size.

The above two components can be predicted in parallel using fixed-size representations of the center points. The architecture of the two heads is shown in Figure \ref{fig:shape_branch}. $P$ represents the image features extracted by the backbone network. Let $F_{shape}\in\mathbb{R}^{H\times W\times S^{2}}$ be the output of the Local Shape head, with $H$ and $W$ represent the height and width of the whole map, and $S^{2}$ represents the number of output channels for this head. The output of the Size head $F_{size}\in\mathbb{R}^{H\times W\times 2}$ is in the same height and width, with a channel size of two. 

For a center point $(x,y)$ at the feature map, the shape feature at this location is extracted by $F_{shape}(x,y)$. The shape vector is in the size of $1\times1\times S^{2}$, and then be reshaped to a 2D shape array of size $S\times S$. 
The size prediction of the center point is $F_{size}(x,y)$, with the height and width being $h$ and $w$. The above 2D shape array is then resized to the size of $h\times w$ to form the final local shape prediction. 

For convenience, the Local Shape Prediction branch is used to refer to the combination of the shape and size heads. This branch produces masks from local point representation, and predicts a local area for each object, which makes it suitable for instance differentiation.

\subsection{Global Saliency Generation}

Although the Local Shape branch generates a mask for each instance, it is not enough for precise segmentation.
As the fixed-size shape vector can only predict a coarse mask, resizing and warping it to the object size losses spatial details, which is a common problem for instance segmentation. Instead of relying on complex pixel calibration mechanism \cite{he2017mask,chen2019tensormask}, we design a simpler and faster approach. 

Motivated by semantic segmentation \cite{long2015fully} which makes pixel-wise predictions on the whole image, we propose to predict a Global Saliency Map to realize pixel level feature alignment. The Map aims to represent the salience of each pixel in the whole image, i.e., whether the pixel belonging to an object area or not. 

Utilizing the fully convolutional backbone, the Global Saliency branch performs the segmentation on the whole image in parallel with the existing Local Shape branch. 
Different from semantic segmentation methods which utilize \textit{softmax} function to realize pixel-wise competition among object classes, our approach uses \textit{sigmoid} function to perform binary classification. The Global Saliency Map can be class-agnostic or class-specific. In the class-agnostic setting, only one binary map is produced to indicate whether the pixels belonging to the foreground or not.  For the class-specific setting, the head produces a binary mask for each object category.

An example of Global Saliency Map is shown in the top of Figure~\ref{fig:network}, using the class-agnostic setting for visualization convenience. As can be seen in the figure, the map highlights the pixels that have saliency, and achieves pixel-wise alignment with the input image.

\subsection{Mask Assembly}\label{sec:asssmbly}

In the end, the Local Shapes and Global Saliency Map are assembled together to form the final instance masks. The Local Shape predicts the coarse area for each instance, and the cropped Saliency Map realizes precise segmentation in the coarse area. Let $L_{k}\in\mathbb{R}^{h\times w}$ represent the Local Shape for one object, and $G_{k}\in\mathbb{R}^{h\times w}$ be the corresponding cropped Saliency Map. They are in the same size of the predicted height and width. 

To construct the final mask, we firstly transform their values to the range of (0,1) using the $sigmoid$ function, then calculate the Hadamard product of the two matrices: 
\begin{equation}
M_{k}=\sigma(L_{k})\odot\sigma(G_{k})
\end{equation}
There is no separate loss for the Local Shape and Global Saliency branch, instead, all supervision comes from the loss function of the assembled mask. Let $T_{k}$ denote the corresponding ground truth mask, the loss function of the final masks is :
\begin{equation}
L_{mask}=\frac{1}{N}\sum_{k=1}^{N} Bce(M_{k},T_{k})
\label{equ:mask}
\end{equation}

where $Bce$ represents the pixel-wise binary cross entropy,
and $N$ is the number of objects.

\subsection{Overall pipeline of CenterMask}

The overall architecture of CenterMask is shown in Figure \ref{fig:network}. 
The Heatmap head is utilized to predict the positions and categories for center points, following a typical keypoint estimation\cite{newell2016stacked} pipeline. Each channel of the output is a heatmap for the corresponding category. Obtaining the center points requires to search the peaks for each heatmap, which are defined as the local maximums within a window. The Offset head is utilized to recover the discretization error caused by the output stride.

Given the predicted center points, the Local Shapes for these points are calculated by the outputs of the Shape head and the Size head at the corresponding locations, following the approach in Section~\ref{sec:local}. The Saliency head produces the Global Saliency Map. In the class-agnostic setting, the output channel number is 1, the Saliency map for each instance is obtained by cropping it with the predicted location and size. In the class-specific setting, the channel of the corresponding predicted category is cropped. The final masks are constructed by assembling the Local Shapes and the Saliency Map.

\textbf{Loss function:}
The overall loss function is composed of four losses: the center point loss, the offset loss, the size loss, and the mask loss. The center point loss is defined in the same way as the Hourglass network \cite{newell2016stacked}, let $\hat{Y}_{ijc}$ be the score at the location (i,j) for class c
in the predicted heatmaps, and $Y$ be the “ground-truth” heatmap. The loss function is a pixel-wise logistic regression modified by the focal loss \cite{lin2017focal}:
\begin{equation}
L_{p}=\frac{-1}{N}\sum_{ijc}\left\{\begin{matrix}
(1-\hat{Y}_{ijc})^{\alpha }\log{(\hat{Y}_{ijc})}\quad \mbox{if}\ Y_{ijc}=1\\ 
(1-Y_{ijc})^{\beta }(\hat{Y}_{ijc})^{\alpha }\log{(1-\hat{Y}_{ijc})}\ \mbox{otherwise}
\end{matrix}\right.
\end{equation}
\noindent where $N$ is the number of center points in the image, $\alpha$ and $\beta$ are the hyper-parameters of the focal loss; The offset loss and size loss follow the same setting of CenterNet \cite{zhou2019objects}, which utilize L1 loss to penalize the distance. Let $\hat{O}$ represent the predicted offset, $p$ represent the ground truth center point, and $R$ represents the output stride, then the low-resolution equivalent of $p$ is $\Tilde{p}=\lfloor \frac{p}{R} \rfloor$, therefore the offset loss is:
\begin{equation}
L_{off}=\frac{1}{N}\sum_{p}\left | \hat{O}_{\Tilde{p}} -(\frac{p}{R}-\Tilde{p})\right |
\end{equation}
Let the true object size be $S_{k}={(h,w)}$, the predicted size be $\hat{S}_{k}={(\hat{h},\hat{w})}$, then the size loss is:
\begin{equation}
L_{size}=\frac{1}{N}\sum_{k=1}^{N}\left | \hat{S}_{k} -S_{k}\right |
\end{equation}
The overall training objective is the combination of the four losses:
\begin{equation}
L_{seg}=\lambda_{p}L_{p}+\lambda_{off}L_{off}+\lambda_{size}L_{size}+\lambda_{mask}L_{mask}
\end{equation}
where the mask loss is defined in Equation~\ref{equ:mask}, $\lambda_{p}$, $\lambda_{off}$, $\lambda_{size}$ and $\lambda_{mask}$ are the coefficients of the four losses respectively.


\subsection{Implementation Details}

\textbf{Train:} Two backbone networks are involved to evaluate the performance of CenterMask: Hourglass-104 \cite{newell2016stacked} and DLA-34 \cite{yu2018deep}. $\rm S$ equals 32 for the shape vector. $\lambda_{p}$, $\lambda_{off}$ and $\lambda_{size}$, $\lambda_{mask}$ are set to 1,1,0.1,1 for the loss function. 
The input resolution is fixed with $512\times512$. All models are trained from scratch, using Adam \cite{article} to optimize the overall objects. The models are trained for 130 epochs, with an initial learning rate of 2.5e-4 and dropped 10$\times$ at 100 and 120 epoch. 
As our approach directly makes use of the same hyper-parameters of CenterNet \cite{zhou2019objects}, we argue that the performance of CenterMask can be improved further if the hyper-parameters are optimized for it correspondingly.

\textbf{Inference:} During testing, 
no data augmentation and no NMS is utilized, only returning the top-100 scoring points with the corresponding masks. The binary threshold for the mask is 0.4.

\begin{table*}[!t]
\begin{subtable}{0.5\linewidth}
\centering
\captionsetup{width=0.95\linewidth}
{\begin{tabular}{l|ccc|ccc}
S& AP & $\rm AP_{50}$ & $\rm AP_{75}$ & $\rm AP_{\textit{S}}$ & $\rm AP_{\textit{M}}$ & $\rm AP_{\textit{L}}$\\
\hline
24&32.0&52.8&33.8&14.0&36.3&48.5\\
32&\textbf{32.5}&\textbf{53.6}&33.9&\textbf{14.3}&36.3&48.7\\
48&\textbf{32.5} &53.4&\textbf{34.1}&13.8&\textbf{36.6}&\textbf{49.0}\\
\end{tabular}}
\caption{\textbf{Size of Shape}: Larger shape size brings more gains.
Performance saturates when S equals 32. Results are based on DLA-34.}\label{tab:1a}
\end{subtable}%
\begin{subtable}{0.5\linewidth}
\centering
\renewcommand\tabcolsep{2.0pt}
\captionsetup{width=0.95\linewidth}
{\begin{tabular}{l|ccc|ccc|c}
Backbone& AP & $\rm AP_{50}$ & $\rm AP_{75}$ & $\rm AP_{\textit{S}}$ & $\rm AP_{\textit{M}}$ & $\rm AP_{\textit{L}}$&FPS\\
\hline
DLA-34&32.5&53.6&33.9&14.3&36.3&48.7&\textbf{25.2}\\
Hourglass-104&\textbf{33.9}&\textbf{55.6}&\textbf{35.5}&\textbf{16.1}&\textbf{37.8}&\textbf{49.2}&12.3\\
\end{tabular}}
\caption{\textbf{Backbone Architecture}: FPS represents frame-per-second. The Hourglass-104 backbone brings 1.4 gains compared with DLA-34, but its speed is more than 2 times slower.}\label{tab:1b}
\end{subtable}%
\vspace*{0.2 cm}
\begin{subtable}{.5\linewidth}
\centering
\captionsetup{width=.95\linewidth}
{\begin{tabular}{l|ccc|ccc}
Shape& AP & $\rm AP_{50}$ & $\rm AP_{75}$ & $\rm AP_{\textit{S}}$ & $\rm AP_{\textit{M}}$ & $\rm AP_{\textit{L}}$\\
\hline
w/o&21.7 &44.7 &18.3 &9.8 &24.0 &31.8\\
w&\textbf{31.5}&\textbf{53.7}&\textbf{32.4}&\textbf{15.1}&\textbf{35.5}&\textbf{45.5}\\
\end{tabular}}
\caption{\textbf{Local Shape branch}: Comparison of CenterMask with or without Local Shape branch.}\label{tab:1c}
\end{subtable}%
\begin{subtable}{.5\linewidth}
\centering
\captionsetup{width=.95\linewidth}
{\begin{tabular}{l|ccc|ccc}
Saliency& AP & $\rm AP_{50}$ & $\rm AP_{75}$ & $\rm AP_{\textit{S}}$ & $\rm AP_{\textit{M}}$ & $\rm AP_{\textit{L}}$\\
\hline
w/o&26.5&51.8&24.5&12.7&29.8&38.2\\
w&\textbf{31.5}&\textbf{53.7}&\textbf{32.4}&\textbf{15.1}&\textbf{35.5}&\textbf{45.5}\\
\end{tabular}}
\caption{\textbf{Global Saliency branch}: Comparison of CenterMask with or without Global Saliency branch. 
}\label{tab:1d}
\end{subtable}%
\vspace*{0.2 cm}
\begin{subtable}{.5\linewidth}\centering
\renewcommand\tabcolsep{3.0pt}
\captionsetup{width=.95\linewidth}
{\begin{tabular}{l|ccc|ccc}
& AP & $\rm AP_{50}$ & $\rm AP_{75}$ & $\rm AP_{\textit{S}}$ & $\rm AP_{\textit{M}}$ & $\rm AP_{\textit{L}}$\\
\hline
Class-Agnostic&31.5&53.7&32.4&15.1&35.5&45.5\\
Class-Specific&\textbf{33.9}&\textbf{55.6}&\textbf{35.5}&\textbf{16.1}&\textbf{37.8}&\textbf{49.2}\\
\end{tabular}}
\caption{\textbf{Class-Agnostic \textit{vs}. Class-Specific}: Comparison of the class-agnostic and class-specific setting of Global Saliency branch. 
}\label{tab:1e}
\end{subtable}
\begin{subtable}{.5\linewidth}\centering
\captionsetup{width=.95\linewidth}
{\begin{tabular}{l|ccc|ccc}
& AP & $\rm AP_{50}$ & $\rm AP_{75}$ & $\rm AP_{\textit{S}}$ & $\rm AP_{\textit{M}}$ & $\rm AP_{\textit{L}}$\\
\hline
w/o&33.9&55.6&35.5&16.1&37.8&49.2\\
w&\textbf{34.4}&\textbf{55.8}&\textbf{36.2}&\textbf{16.1}&\textbf{38.3}&\textbf{50.2}\\
\end{tabular}}
\caption{\textbf{Direct Saliency supervision}: Comparison of CenterMask with or without direct Saliency supervision. 
}\label{tab:1f}

\end{subtable}
\vspace{-0.2cm}
\caption{Ablation experiments of CenterMask. All models are trained on \texttt{trainval35k} and tested on \texttt{minival}, using the Hourglass-104 backbone unless otherwise noted.}\label{tab:1}
\end{table*}

\section{Experiments}
The performance of the proposed CenterMask is evaluated on the MS COCO instance segmentation benchmark \cite{lin2014microsoft}. The model is trained on the ~115k \texttt{trainval35k} images and tested on the 5k \texttt{minival} images. Final results are evaluated on 20k \texttt{test-dev}. 

\subsection{Ablation Study}

\begin{figure*}
    \begin{minipage}{0.36\linewidth}
    \begin{subfigure}{\linewidth}
    \includegraphics[width=\linewidth]{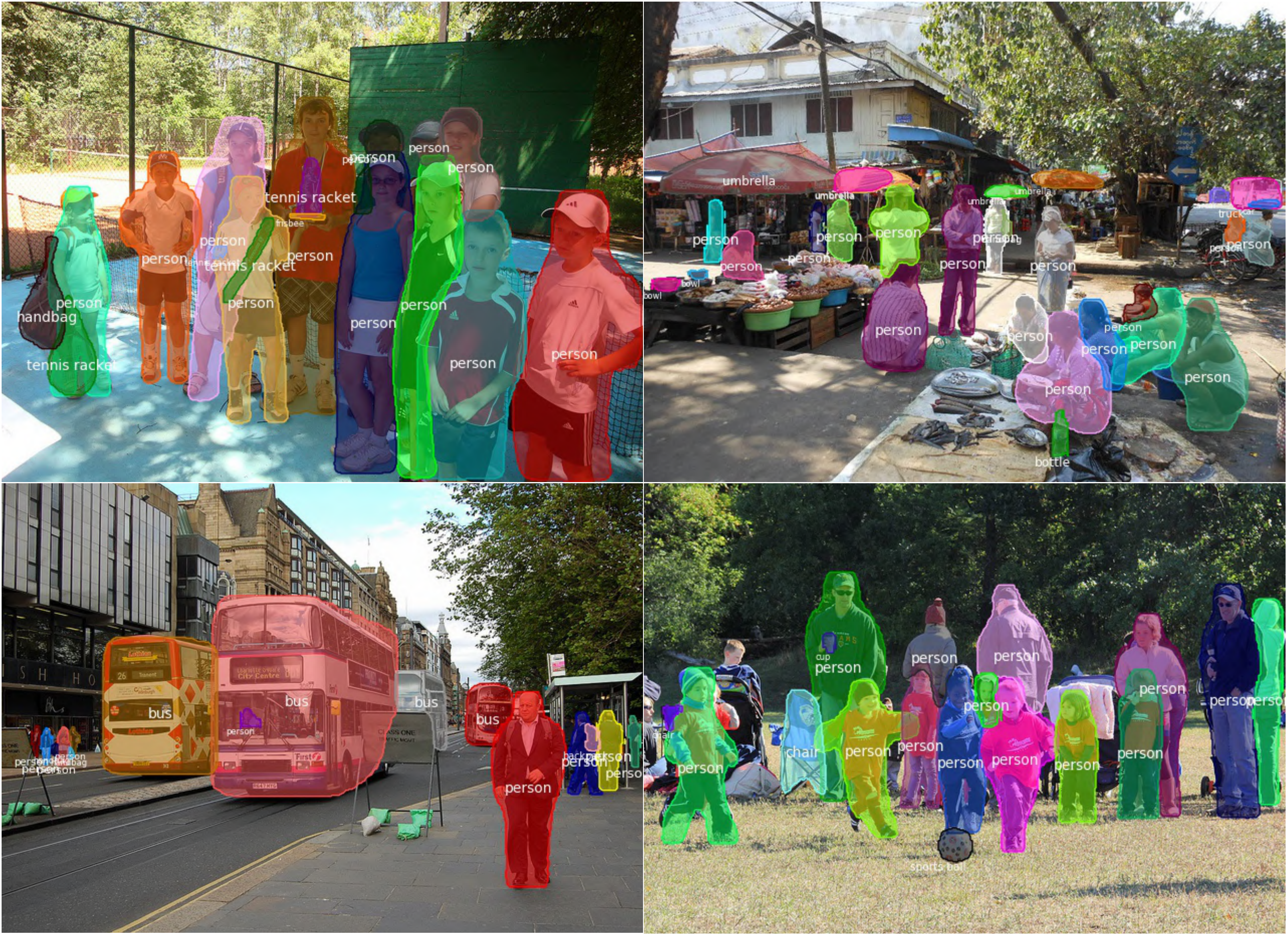}
    \caption{\textbf{Results of CenterMask in Shape-only setting.} The Local Shape branch seperates instances with coarse masks.}
    \label{fig:shape-only}
    \end{subfigure}
    \begin{subfigure}{\textwidth}
    \includegraphics[width=\linewidth]{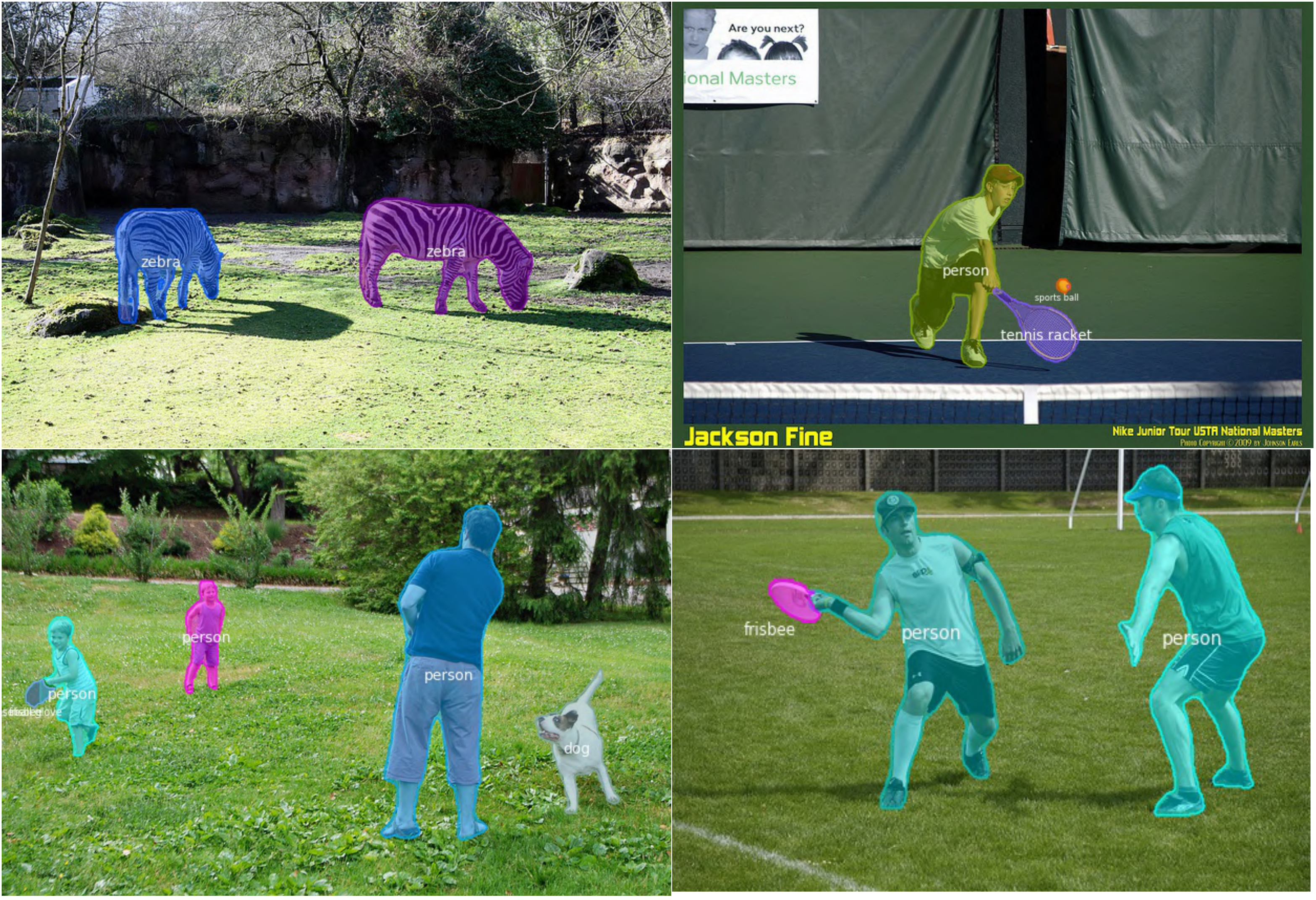}
    \caption{\textbf{Results of CenterMask in Saliency-only setting.} The Global Saliency branch performs well when there are no overlap between objects.}
    \label{fig:saliency-only}
    \end{subfigure}
    \end{minipage}
    \hfil
    \begin{minipage}{0.62\linewidth}
    \begin{subfigure}{\linewidth}
    \includegraphics[width=\linewidth]{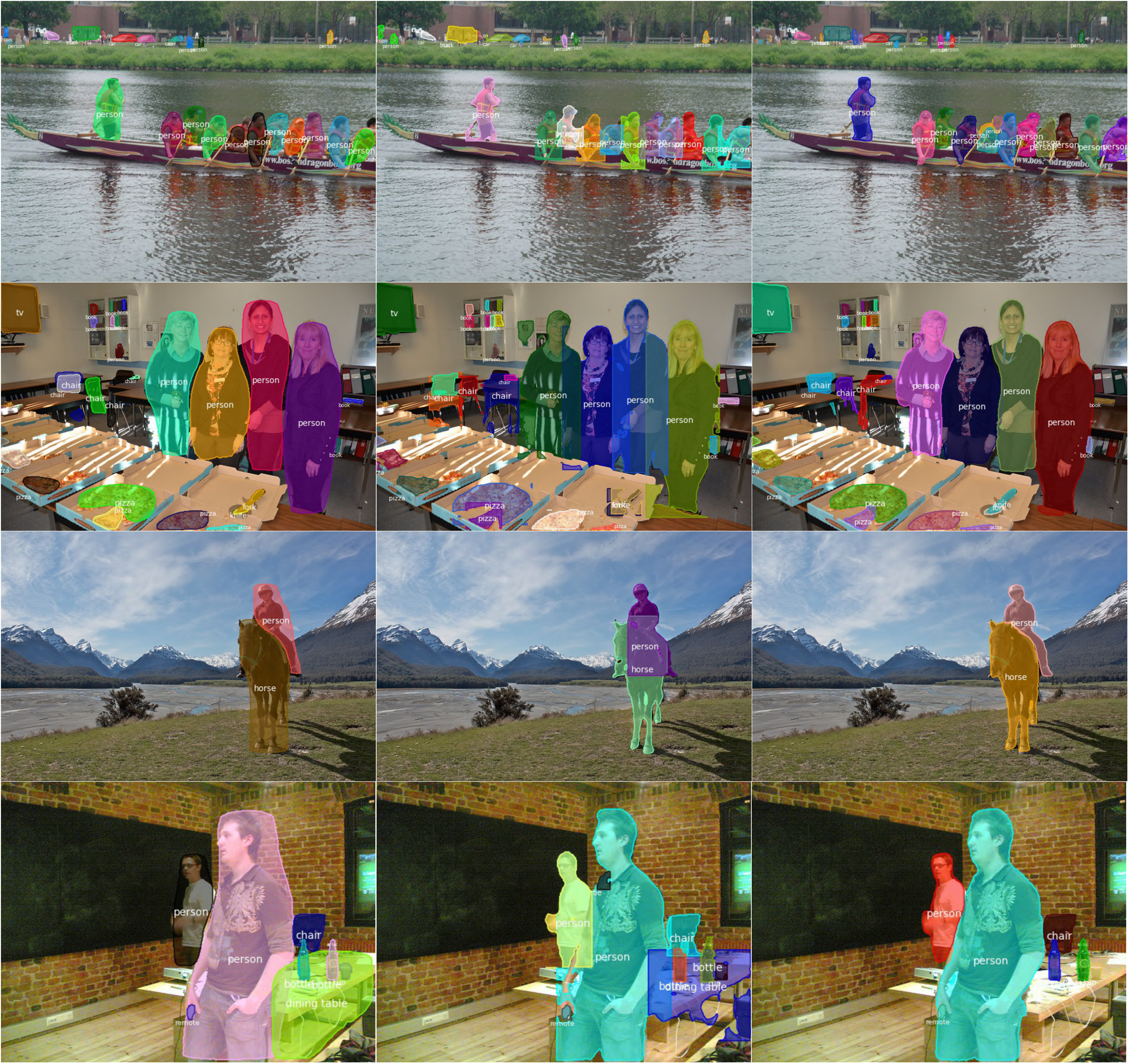}
    \caption{\textbf{Comparison of CenterMask results in challenging conditions}. Images form left to right are generated by: Shape-only, Saliency-only and the combination of the two branches. }
    \label{fig:combine}
    \end{subfigure}
    \end{minipage}
\caption{\textbf{Images generated by CenterMask in different settings.} The Saliency branch is in class-agnostic setting for this experiment.}
\label{fig:Ablation}
\vspace{-0.2cm}
\end{figure*}

\begin{table*}[t]
\begin{center}
\begin{tabular}{l|c|c|c|ccc|ccc}

Method & Backbone & Resolution & FPS & AP & $\rm AP_{50}$ & $\rm AP_{75}$ & $\rm AP_{\textit{S}}$ & $\rm AP_{\textit{M}}$ & $\rm AP_{\textit{L}}$\\
\hline
\textit{two-stage}&&&&&&\\
MNC \cite{dai2016instance2} &ResNet-101-C4 &-&2.78 &24.6 &44.3 &24.8 &4.7 &25.9 &43.6\\
FCIS \cite{li2017fully} &ResNet-101-C5-dilated &multi-scale&4.17 &29.2 &49.5 &- &7.1 &31.3 &50.0\\
Mask R-CNN \cite{he2017mask}&ResNeXt-101-FPN &800$\times$1333&8.3 &37.1 &60.0 &39.4 &16.9 &39.9 &53.5\\
\hline
\textit{one-stage}&&&&&&\\
ExtremeNet \cite{zhou2019bottomup}&Hourglass-104&512$\times$512&3.1&18.9&44.5&13.7&10.4&20.4&28.3\\
TensorMask \cite{chen2019tensormask}&ResNet-101-FPN&800$\times$1333&2.63&37.3&59.5&39.5&17.5&39.3&51.6\\
YOLACT \cite{bolya-iccv2019}&ResNet-101-FPN&700$\times$700&23.6&31.2&50.6&32.8&12.1&33.3&47.1\\
YOLACT-550 \cite{bolya-iccv2019}&ResNet-101-FPN&550$\times$550&33.5&29.8&48.5&31.2&9.9&31.3&47.7\\
PolarMask \cite{xie2019polarmask}&ResNeXt-101-FPN&768$\times$1280& 10.9&32.9&55.4&33.8&15.5&35.1&46.3\\
\textbf{CenterMask}&DLA-34&512$\times$512&25.2&33.1&53.8&34.9&13.4&35.7&48.8\\
\textbf{CenterMask}&Hourglass-104&512$\times$512&12.3&34.5&56.1&36.3&16.3&37.4&48.4\\
\end{tabular}
\end{center}
\vspace{-0.5cm}
\caption{\textbf{Instance segmentation mask AP on COCO \texttt{test-dev}.} Resolution represents the image size of training. We show single scale testing for most models. Frame-per-second (FPS) were measured on the same machine whenever possible. A dash indicates the data is not available. }
\label{tab:2}
\end{table*}

A number of ablation experiments are performed to analyze CenterMask. Results are shown in Table \ref{tab:1}.

\textbf{Shape size Selection:} Firstly, the sensitivity of our approach to the size of the Local Shape representation is analyzed in Table~\ref{tab:1a}. Larger shape size brings more gains, but the difference is not large, indicating that the Local Shape representation is robust to the feature size. When S equals 32, the performance saturates, therefore we use the number as the default Shape size. 

\textbf{Backbone Architecture:} Results of CenterMask with different backbones are shown in Table~\ref{tab:1b}. The large Hourglass brings about 1.4 gains compared with the smaller DLA-34 \cite{yu2018deep}. The model with DLA-34 \cite{yu2018deep} backbone realizes 32.5 mAP with 25.2 FPS, achieving a good speed-accuracy trade-off. 

\textbf{Local Shape branch:} 
The comparison of CenterMask with or without Local Shape branch is shown in Table~\ref{tab:1c}, with Saliency branch in class-agnostic setting. The Shape branch brings about 10 gains. Moreover, CenterMask with only the Shape branch achieves 26.5 AP (as shown in the first row of Table~\ref{tab:1d}), images generated by this model are shown in Figure~\ref{fig:shape-only}. Each image contains multiple objects with dense overlaps, the Shape branch can separate them well with coarse masks. The above results illustrate the effectiveness of the proposed Local Shape branch.

\textbf{Global Saliency branch:}
The comparison of CenterMask with or without Global Saliency branch is shown in Table~\ref{tab:1d}, introduction of the Saliency branch improves 5 points, compared with model with only Local Shape branch. 

We also conduct visualization to CenterMask with only Saliency branch. As shown in Figure~\ref{fig:saliency-only}, there is no overlap between objects in these images. The Saliency branch performs good enough for this kind of situation by predicting precise mask for each instance, indicating the effectiveness of this branch for pixel-wise alignment.  

Moreover, the two settings of the Global Saliency branch are compared in Table~\ref{tab:1e}. The class-specific setting achieves 2.4 points higher than the class-agnostic counterpart, showing that the class-specific setting can help separate instances from different categories better.

For the class-specific version of Global Saliency branch, a binary cross-entropy loss is added to supervise the branch directly besides the mask loss Eq.~(\ref{equ:mask}). The comparison of CenterMask with or without the new loss is shown in Table~\ref{tab:1f}, direct supervision brings 0.5 points.

\textbf{Combination of Local Shape and Global Saliency:} Although the Saliency branch performs well in non-overlapping situations, it can not handle more complex images. We conduct the comparison of Shape-only, Saliency-only and the Combination of both in challenging conditions of instance segmentation. As shown in Figure~\ref{fig:combine}, objects overlap exists in these images. In the first column, the Shape branch separates different instances well, but the predicted masks are coarse. In the second column, the Saliency branch realizes precise segmentation but fails in the overlapping situations, which results in obvious artifacts on the overlapping area. CenterMask with both branches inherits their merits and avoid their weakness. As shown in the last column, overlapped objects are separated well and segmented precisely simultaneously, illustrating the effectiveness of our proposed model.

\subsection{Comparison with state-of-the-art}

\begin{figure*}[!t]
\centering
\includegraphics[width=.95\linewidth]{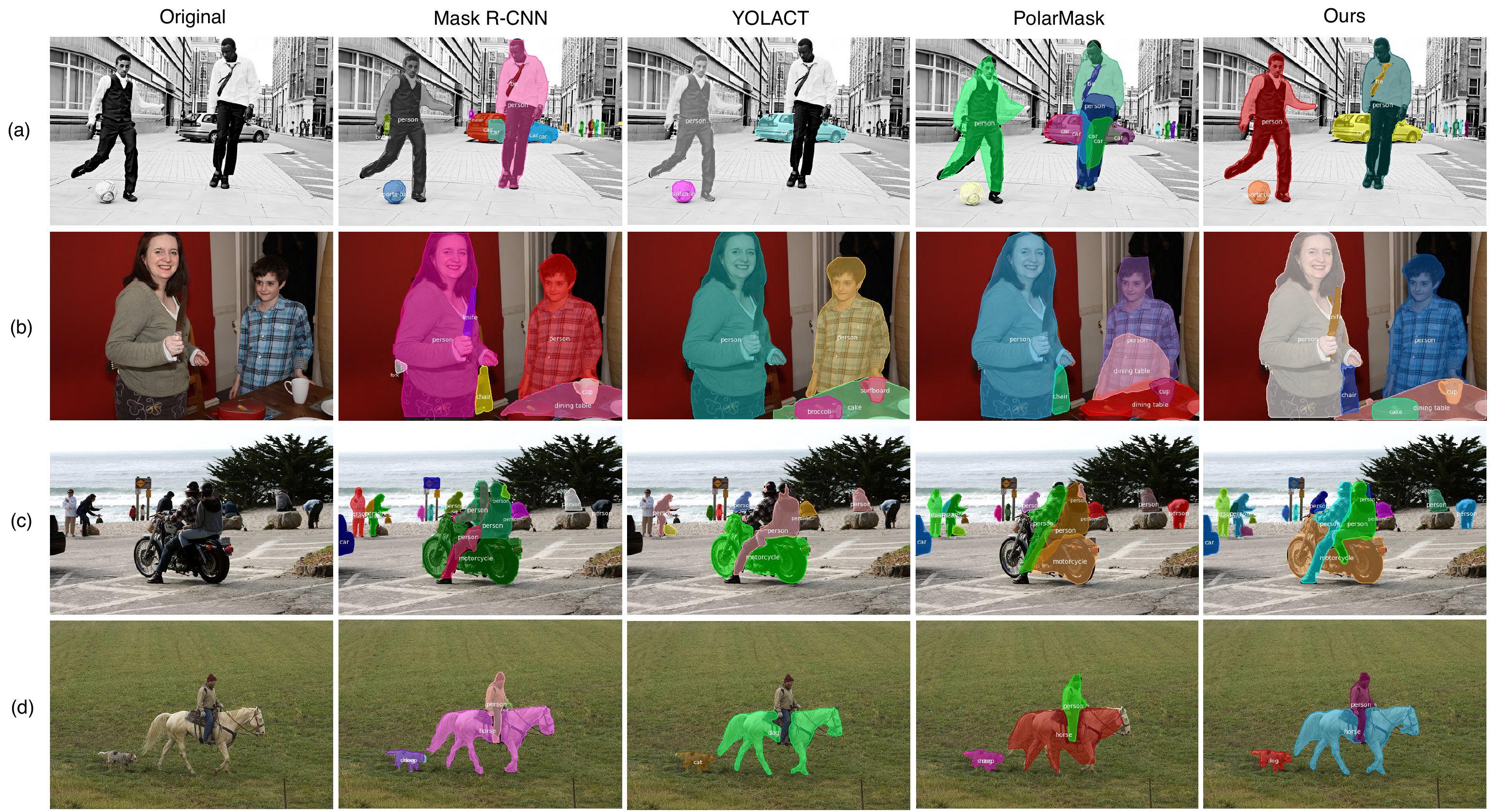}
\vspace{-0.2cm}
\caption{\textbf{Visualization comparison of three different instance segmentation methods.} From left to right are the results of : Original image, Mask R-CNN, YOLACT, PolarMask, and our method on COCO \texttt{minival} images. 
}
\label{fig:compare}
\vspace{-0.3cm}
\end{figure*}

In this section, we compare CenterMask with the state-of-the-art instance segmentation methods on the COCO\cite{lin2014microsoft} \texttt{test-dev} set. 

As a one-stage instance segmentation method, our model follows a simple setting to perform the comparison: totally trained from scratch without pre-trained weights\cite{deng2009imagenet} for the backbone, using a single model with single-scale training and testing, and inference without any NMS.

As shown in Table~\ref{tab:2}, two models achieve higher AP than our method: the two-stage Mask R-CNN and the one-stage TensorMask, but their speed is 4 fps and 5 times slower than our largest model respectively. We think the gaps arise from the complicated and time-consuming feature align operations. Compared with the most accurate model of YOLACT \cite{bolya-iccv2019}, CenterMask with DLA-34 backbone achieves a higher AP with a faster speed. Compared with PolarMask \cite{xie2019polarmask}, CenterMask with hourglass-104 backbone is 1.6 point higher with a faster speed. 

Figure~\ref{fig:compare} shows the visualization of the results generated by the state-of-the-art models, only comparing the ones that have released code. Mask R-CNN \cite{he2017mask} detects objects well, but there are still artifacts in the masks, such as the heads of the two people in (a), we suppose it is caused by feature pooling. The YOLACT \cite{bolya-iccv2019} segments instance precisely, but misses object in (d) and fails in some overlapping situations, such as the two legs in (c). The PolarMask can separate different instances, but its mask is not precise due to the polygon mask representation. Our CenterMask can separate overlapping objects well and segment masks precisely.

\subsection{CenterMask on FCOS Detector}

Besides CenterNet\cite{zhou2019objects}, the proposed Local Shape and Global Saliency branches can be embedded into other off-the-shelf detection models easily. FCOS\cite{tian2019fcos}, which is one of the state-of-the-art one stage object detectors, is utilized to perform the experiment. The performance of CenterMask built on FCOS with different backbones are shown in Table~\ref{tab:fcos}, with the training followings the same setting of Mask R-CNN\cite{he2017mask}. With the same backbone of ResNeXt-101-FPN, CenterMask-FCOS achieves 3.8 points higher than PolarMask\cite{xie2019polarmask} in Table~\ref{tab:2}, and the best model achieves 38.5 mAP on COCO test-dev, showing the generalization of CenterMask.

To show the superiority of CenterMask on precise segmentation, we evaluate the model on the higher-quality LVIS annotations. The results are shown in Table ~\ref{tab:lvis}. Based on the same backbone, the CenterMask-FCOS achieves better performance than Mask R-CNN.

\begin{table}
\begin{center}
\renewcommand\tabcolsep{1.3pt}
\begin{tabular}{l|ccc|ccc}
Backbone & AP & $\rm AP_{50}$ & $\rm AP_{75}$ & $\rm AP_{\textit{S}}$ & $\rm AP_{\textit{M}}$ & $\rm AP_{\textit{L}}$\\
\hline
ResNet-101-FPN&36.1&58.7&38.0&16.5&38.4&51.2\\
ResNeXt-101-FPN&36.7&59.3&38.8&17.4&38.7&51.4\\
ResNet-101-FPN-DCN&37.6&60.4&39.8&17.3&39.8&53.4\\
ResNeXt-101-FPN-DCN&38.5&61.5&41.0&18.7&40.5&54.8
\end{tabular}
\end{center}
\vspace{-0.65cm}
\caption{\textbf{Performance of CenterMask-FCOS on COCO test-dev.} DCN represents deformable convolution\cite{Dai_2017_ICCV}.}
\label{tab:fcos}
\end{table}

\begin{table}[t]
\begin{center}
\renewcommand\tabcolsep{9.0pt}
\begin{tabular}{l|c|c}
Model & Backbone & AP \\
\hline
Mask R-CNN\cite{he2017mask}&ResNet-101-FPN&36.0\\
CenterMask-FCOS&ResNet-101-FPN&40.0\\
\end{tabular}
\end{center}
\vspace{-0.65cm}
\caption{\textbf{Performance of CenterMask-FCOS on LVIS\cite{gupta2019lvis}.} The AP of Mask R-CNN comes from the original LVIS paper.}
\label{tab:lvis}
\vspace{-0.3cm}
\end{table}

\section{Conclusion}
In this paper, we propose a single shot and anchor-box free instance segmentation method, which is simple, fast and accurate. The mask prediction is decoupled into two critical modules: the Local Shape branch to separate different instances effectively and the Global Saliency branch to realize precise segmentation pixel-wisely. Extensive ablation experiments and visualization images show the 
effectiveness of the proposed CenterMask. We hope our work can help ease more instance-level recognition tasks.

\noindent\textbf{Acknowledgements}
This research is supported by Beijing Science and Technology Project (No. Z181100008918018).

{\small
\bibliographystyle{ieee_fullname}
\bibliography{egbib}
}

\end{document}